\documentclass[10pt,twocolumn,letterpaper]{article}

\usepackage{cvpr}
\usepackage{times}
\usepackage{epsfig}
\usepackage{graphicx}
\usepackage{amsmath}
\usepackage{amssymb}
\usepackage{bm}
\usepackage{tabularx}

\usepackage{algorithm}
\usepackage{algpseudocode}

\makeatletter
\def\BState{\State\hskip-\ALG@thistlm}
\makeatother

\algnewcommand\algorithmicinput{\textbf{INPUT:}}
\algnewcommand\INPUT{\item[\algorithmicinput]}

\renewcommand{\vec}[1]{\text{\boldmath$#1$}} 

\DeclareMathOperator*{\argmax}{\arg\!\max}

\algnewcommand\algorithmicoutput{\textbf{OUTPUT:}}
\algnewcommand\OUTPUT{\item[\algorithmicoutput]}

\usepackage[pagebackref=true,breaklinks=true,letterpaper=true,colorlinks,bookmarks=false]{hyperref}

 \cvprfinalcopy % *** Uncomment this line for the final submission

\def\cvprPaperID{4277} % *** Enter the CVPR Paper ID here

\newcommand{\fm}[1]{{\textcolor[rgb]{1,0,0}{#1}}}

\newcommand{\R}{\mathbb{R}}
\newcommand{\N}{\mathcal{N}}
\newcommand{\Sp}{\mathcal{S}}
\newcommand{\G}{\mathcal{G}}
\newcommand{\T}{\mathcal{T}}
\newcommand{\D}{\mathcal{D}}
\newcommand{\Lno}{\mathcal{L}}

\newcommand{\X}{\mathcal{X}}

\newcommand\norm[1]{\left\lVert#1\right\rVert}

\makeatletter
\newcommand{\multiline}[1]{%
  \begin{tabularx}{\dimexpr\linewidth-\ALG@thistlm}[t]{@{}X@{}}
    #1
  \end{tabularx}
}
\makeatother

\ifcvprfinal\fi
\begin{document}

%%%%%%%%% TITLE
\title{Unsupervised Reverse Domain Adaptation for Synthetic Medical Images via Adversarial Training}

\author{Faisal Mahmood$^1$  \quad   Richard Chen$^2$  \quad    Nicholas J. Durr$^1$\\
$^1$Department of Biomedical Engineering  \quad    $^2$Department of Computer Science\\
Johns Hopkins University (JHU)\\
{ \{faisalm, rchen40, ndurr\}@jhu.edu}
% For a paper whose authors are all at the same institution,
% omit the following lines up until the closing ``}''.
% Additional authors and addresses can be added with ``\and'',
% just like the second author.
% To save space, use either the email address or home page, not both
%%%%%{\tt\small secondauthor@i2.org}
}

\maketitle
%\thispagestyle{empty}

%%%%%%%%% ABSTRACT
\begin{abstract}

To realize the full potential of deep learning for medical imaging, large annotated datasets are required for training. Such datasets are difficult to acquire because labeled medical images are not usually available due to privacy issues, lack of experts available for annotation, underrepresentation of rare conditions and poor standardization. Lack of annotated data has been addressed in conventional vision applications using synthetic images refined via unsupervised adversarial training to look like real images. However, this approach is difficult to extend to general medical imaging because of the complex and diverse set of features found in real human tissues. We propose an alternative framework that uses a reverse flow, where adversarial training is used to make real medical images more like synthetic images, and hypothesize that clinically-relevant features can be preserved via self-regularization. These domain-adapted images can then be accurately interpreted by networks trained on large datasets of synthetic medical images. We test this approach for the notoriously difficult task of depth-estimation from endoscopy. We train a depth estimator on a large dataset of synthetic images generated using an accurate forward model of an endoscope and an anatomically-realistic colon. This network predicts significantly better depths when using synthetic-like domain-adapted images compared to the real images, confirming that the clinically-relevant features of depth are preserved. 

%This network, which uses a domain-adapted synthetic-like representation of real images, predicts significantly better depths as compared to the state-of-the-art using the real endoscopy images, confirming that the clinically-relevant features of depth are preserved. 

%When applied to domain-adapted synthetic-like input images, this network performs significantly better than 

%and hypothesize that diagnostic features from the real image can be preserved via self-regularization
\end{abstract}

%%%%%%%%% BODY TEXT
\section{Introduction}

\begin{figure}
\label{fig_1}
\centering
\includegraphics[width=8.5cm]{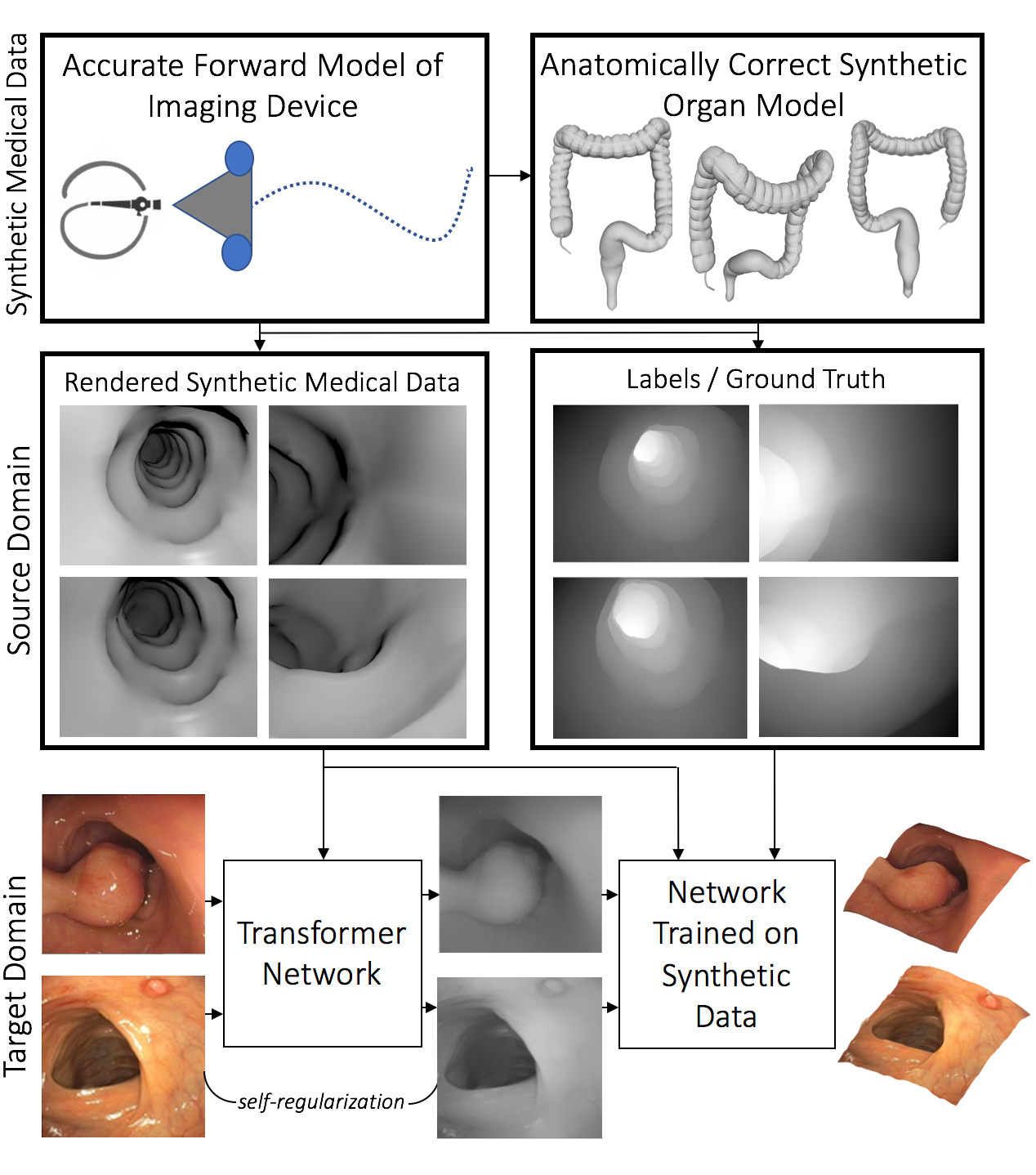}
\caption{Unsupervised reverse domain adaption for endoscopy images. We use an accurate forward model of an endoscope and an anatomically correct colon model to generate synthetic endoscopy images with ground truth depth. This large synthetic dataset can be used to train a deep network for depth estimation. An adversarial network transforms input endoscopy images to a synthetic-like representation while preserving clinically relevant features via self-regularization. These synthetic-like images can be directly used for depth estimation from the network trained on synthetic images.}
\end{figure}

%- Importance of labeled datasets for medical imaging.

%Machine learning and biomedical imaging are a match made in heaven

Deep Learning offers great promise for the reconstruction and interpretation of medical images \cite{shen2017deep,greenspan2016guest}. Countless applications in clinical diagnostics, disease screening, interventional planning, and therapeutic surveillance rely on the subjective interpretation of medical images from health-care providers. This approach is costly, time-intensive, and has well-known accuracy and precision limitations\textemdash all of which could be mitigated by objective, automatic image analysis.

For conventional images, deep learning has achieved remarkable performance for a variety of computer vision tasks, typically by utilizing large sets of real-world images for training, such as ImageNet \cite{krizhevsky2012imagenet}, COCO \cite{lin2014microsoft} and Pascal VOC \cite{everingham2015pascal}. Unfortunately, the potential benefits of deep learning have yet to transfer to the most critical needs in medical imaging because there are no large, annotated dataset of medical images available. Despite the compelling need for such a dataset, there are practical concerns that impede its development, including the cost, time, expertise, privacy, and regulatory issues associated with medical data collection, annotation, and dissemination.

%Applying modern deep-learning enabled machine learning methods to medical imaging offers great promise for a variety of applications including both image reconstruction and interpretation \cite{shen2017deep,greenspan2016guest}. Most current medical image interpretations are preformed by humans and vast inconsistencies and variations exist across interpreters. It has been demonstrated that deep-learning can be a key enabler for improving, automating and standardizing medical image reconstruction and analysis. However, deep learning methods rely on the availability of large scale databases such as ImageNet \cite{krizhevsky2012imagenet}, COCO \cite{lin2014microsoft} and Pascal VOC \cite{everingham2015pascal} for real world images. Collecting labeled and annotated data is difficult and expensive for conventional vision applications but it becomes exponentially tedious in the medical imaging domain. 

%Major limiting factors in collecting large amounts of ground truth and labeled medical data include: a) the cost and time associated with medical data collection which often involves deviation from standard medical protocol and requires regulatory approvals b) privacy issues c) non availability of experts for labeling thousands of images d) underrepresentation of rare conditions in the available data. 

The obstacles associated with developing a large dataset of real images can be circumvented by generating synthetic images \cite{qiu2016unrealcv,shafaei2016play,rusu2016sim}. Considerable effort has been devoted to adapting models generated with synthetic data as the source domain to real data as the target domain \cite{bousmalis2016unsupervised}. Advances in adversarial training have sparked interest in making synthetic data look more realistic via unsupervised adversarial learning (SimGAN) \cite{shrivastava2016learning}. In the medical imaging domain, there has been recent success in generating realistic synthetic data for the relatively constrained problem of 2D retinal imaging using standard GANs \cite{guibas2017synthetic}. In more complex applications, it is challenging to generate an appropriate span of synthetic medical images for training, because few models exist that accurately simulate the anatomical complexity and diversity found in healthy to pathologic tissues. Moreover, the forward models for medical imaging devices are more complex than those used in many conventional vision applications. Consequently, models trained on synthetic medical data may fail to generalize to real medical images, where accurate interpretation may be critically important.

Cross-patient network usage is a well-known challenge to learning-based medical imaging methods. Often a network trained on data from one patient fails to generalize to other patients. This is commonly observed for optical imaging methods, such as endoscopy, which capture both low- and high-level texture details of the patient. Low-level texture details are patient-specific and not diagnostic, such as vascular patterns. High-level texture, on the other hand, contains clinically-relevant features that should be generalized across patients. This complication makes it difficult for methods like SimGAN \cite{shrivastava2016learning} to work both accurately and generally because the span of realistic images produced will be similar to the real images used for training.

In this work, we propose to reverse the flow of traditional adversarial training-based domain adaption.  Instead of changing synthetic images to appear realistic \cite{shrivastava2016learning}, we transform real images to look more synthetic (Fig. 1). We train an adversarial transformation network which transforms real medical images to a synthetic-like representation while preserving clinically-relevant information. In summary, we can train solely on synthetic medical data as the source domain and transform real data in the target domain to a more synthetic interpretation, thus bridging the gap between the source and target domains in a reverse manner.

To transform real images to a synthetic-like representation, we train a transformer with an adversarial loss similar to GANs \cite{goodfellow2014generative} and SimGAN \cite{shrivastava2016learning}. However, unlike SimGAN that trains for inducing realism to synthetic data, we train for a synthetic-like representation of real data. With the roles of synthetic and real data reversed, the overall transformer architecture is similar to a standard GAN and is composed of a transformer network that tries to fool a discriminator network into thinking that the transformed medical image is synthetic. In addition to removing patient specific details from the data, the synthetic image should preserve enough information within the data that it could be used for the task at hand. To preserve this information a fully connected network is used and the adversarial loss is complemented with a self-regularization term which constrains the amount of deviation from the real image.\\

%With the roles of synthetic and real data reversed the overall transformer architecture is similar to a standard GAN 

%which is composed of a generator network that attempts to make synthetic medical images that are indistinguishable from real images. Additionally, removing patient specific details from the data the generated 'synthetic-like' image should preserve enough information within the data that it could be used for the task at hand. To preserve this information, a fully-connected network is used and the adversarial loss is complemented with a self-regularization term which constrains the amount of deviation from the real image.\\

\textbf{Contributions}
\vspace{-0.7em}
\begin{enumerate}
\item We propose an adversarial training-based reverse domain adaptation method which uses unlabeled synthetic data to transform real data to a synthetic-like representation while maintaining clinically relevant diagnostic features via self-regularization.
\vspace{-0.7em}
\item \textbf{Synthetic Endoscopy Data Generation:} We generate a large dataset of perfectly-annotated synthetic endoscopy images from an endoscope forward model and an anatomically correct colon model.
\vspace{-0.7em}
\item  \textbf{Reverse Domain Adaptation:} We train a transformer network via adversarial training composed of a generator which generates the a synthetic-like representation of real endoscopy images. The loss function in the generator contains a discriminator to classify the endoscopy images as real or synthetic and a self-regularization term that penalizes large deviations from the real image.
\vspace{-0.7em}
\item \textbf{Qualitative and Quantitative Study:} We validate our domain adaptation approach by using synthetically generated endoscopy data to train a monocular endoscopy depth estimation network and quantitatively testing it with real endoscopy data from: a) Colon Phantom b) Porcine Colon and qualitatively testing it with real human endoscopy data. We further show that the depth obtained from training on synthetic data can be used to improve state-of-the-art results on polyp segmentation.
%\vspace{-0.7em}
%\item To further validate the effectiveness of our approach we validate the depth estimation network with real endoscopy data collected from a porcine colon and its ground true depth acquired from a CT experiment.

\end{enumerate}

%Fig. 1 explains this flow for endoscopy images. Learning from monocular endoscopy images is a notoriously difficult task because there is no ground truth data available since it is not possible to incorporate a depth sensor on an endoscope due to size limitations and regularity restrictions. Thus we train a depth estimation model using synthetic endoscopy data. We then transform the data 

%

%\textbf{Contributions}{}

\section{Related Work}

\textbf{Navigating Limited Medical Imaging Data:} Improving the performance of deep learning methods with limited data is an active research area. Standard data augmentation has been used for medical imaging for the past years. Ronneberger \textit{et al.} \cite{ronneberger2015u} demonstrated success with using elastic augmentation with U-Net architectures for medical image segmentation. Payer \textit{et al.} \cite{payer2016regressing} have demonstrated incorporating application specific \textit{a priori} information can train better deep networks. There is a growing interest in transferring knowledge from networks trained for conventional vision to the medical imaging domain \cite{zhu2016generative}. However, the major limiting factor with all these approaches is the fact that there is very limited data to train from.

\textbf{Generative Adversarial Networks:} The GAN framework was first presented by Goodfellow \textit{et. al.} in \cite{goodfellow2014generative} and was based on the idea of training two networks, a generator and a discriminator simultaneously with competing losses. While the generator learns to generate realistic data from a random vector, the discriminator classifies the generated image as real or fake and gives feedback to the generator. Once the training reaches equilibrium the generator is able to fool the discriminator every time it generates a new image. Initially GANs were applied to the MINST dataset \cite{goodfellow2014generative} but recently the framework has been refined and used for a variety of applications \cite{salimans2016improved}. Models with adversarial losses have been used for synthesis of 3D shapes, image-to-image translation, for generating radiation patterns etc. Recently, Zhu \textit{et al.} \cite{zhu2016generative} proposed iGAN which enables interactive image manipulation on a natural image manifold. Shrivistava \textit{et al.} \cite{shrivastava2016learning} have proposed an unsupervised method for refining synthetic images to look more realistic using a modified adversarial training framework.

\textbf{Adversarial Training for Biomedical Imaging:} Various kinds of adversarial training has recently been used for a variety of medical imaging tasks including noise reduction \cite{wolterink2017generative}, segmentation \cite{zhang2017deep,moeskops2017adversarial}, detection \cite{kohl2017adversarial}, reconstruction \cite{mardani2017deep}, classification \cite{zhang2017semi} and image synthesis \cite{osokin2017biogans}. Osokin \textit{et al.} \cite{osokin2017biogans} use GANs for synthesizing biological cells imaged by fluorescence microscopy. Costa \textit{et  al.} \cite{costa2017end} and Guibas \textit{et al.} \cite{guibas2017synthetic} synthesize retinal images using adversarial training. All current adversarial training-based image synthesis methods attempt to generate realistic images from a random noise vector or refine synthetic images to create more realistic images, in contrast our method, transforms real images to a synthetic-like representation allowing the desired network to be trained only on synthetic images.

%\textbf{Monocular Endoscopy Depth Estimation:} 

\section{Generating Synthetic Medical Data}

Despite the widespread use of synthetic data for training deep networks for real world images \cite{su2015render,gupta2016synthetic,varol2017learning,planche2017depthsynth}, its use for medical imaging applications has been relatively limited. Unlike conventional real-world images that may contain a constrained span of object diversity, medical images capture information of biological tissues which contain unique patient-specific texture that is difficult to model. We therefore propose a frame work where we generate a large dataset of medical images with this patient-specific detail removed so that a network can be trained on universal diagnostic features. In general, this synthetic data can be generated by (Fig. 2):

\begin{enumerate}
\item Developing an accurate forward model for the medical imaging device. 
\item Generating an anatomically accurate model of the organ being imaged.
\item Rendering images from a variety of positions, angles and parameters.
\end{enumerate}

Typically, forward models for medical imaging devices is more complicated as compared to typical cameras, and anatomically accurate models need to represent a high degree of variation and rare conditions to cater for a diverse set of patients. 

\textbf{Synthetic Endoscopy Data with Ground Truth Depth:} For the purpose of demonstration of our proposed methods we focus on the task of depth estimation from monocular endoscopy images. This is a notoriously difficult problem because of the lack of clinical images with available ground truth data, since it is difficult to include a depth sensor on an endoscope. We generate synthetic data to overcome this issue. We develop a forward model of an endoscope with a wide-angle monocular camera and two to three light sources that exhibit realistic inverse square law intensity fall-off. We use a synthetically generated and anatomically accurate colon model and image it using the virtual endoscope placed at a variety of angles and varying conditions to mimic the movement of an actual endoscope. We also generate pixel-wise ground truth depth for each rendered image. We finally create a dataset with 260,000 images with ground truth depth. Although this large dataset of images is able to train efficient deep networks these networks are not effectively generalizable to real world images.

\begin{figure}
\label{fig_2}
\centering
\includegraphics[width=8.5cm]{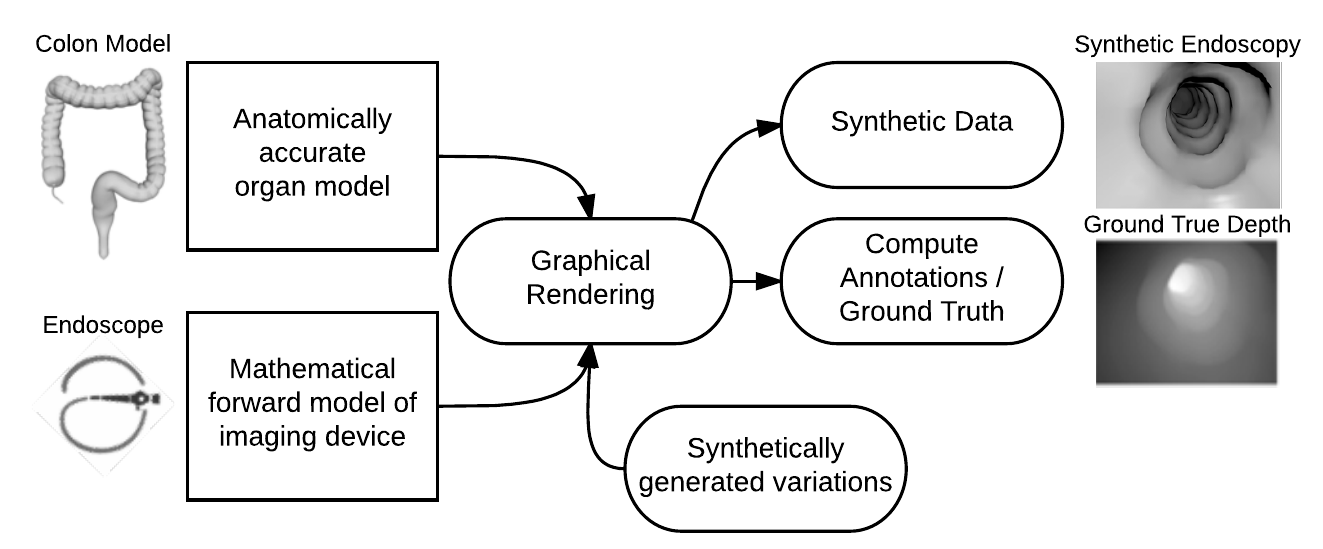}
\caption{General framework for generating synthetic medical imaging data with endoscopy as an example.}
\end{figure}

\section{Reverse Domain Adaptation}

%\textbf{Objective Function:}

\textbf{Transformer Loss:}  Formally, the goal of our proposed reverse domain adaptation method is to use a set of synthetic images $\bm{g}_i\in\G$ to learn a transformer $\bm{x}^{'}=\T_{\gamma_{t}}(\bm{x})$ that can transform real images $\bm{x}$ to a synthetic-like representation $\bm{x}^{'}$. The transformer should be able to fool a discriminator $\D_{\gamma_{d}}$ where $\gamma_t$ and $\gamma_d$ are the learning parameters. There are three key requirements for this setup: a) The transformer output should only remove the patient specific details in the image, while preserving diagnostic features. b) The adversarial training should not introduce artifacts in the transformed image. c) The adversarial training should be stable. The transformer loss function can be defined as,

\begin{equation}\label{eq:Ttans_loss}
  \begin{aligned}
 \Lno_{\T}(\gamma_t) = \sum_{i} \psi(\bm{x}_i,\G;{\gamma_t}) + \lambda \phi(\bm{x}_i;\gamma_t),
  \end{aligned}
\end{equation}

where, $\psi$ forces the real image to a synthetic-like representation and $\phi$ penalizes large variations to preserve specific properties of the real image. $\lambda$ controls the amount of self-regularization enforced by $\phi$. 

\textbf{Discriminator Loss:} In order to transform a real image to its synthetic-like counterpart, the gap between the representations of the real and synthetic image needs to be minimized. An ideal transformer should be able to produce an indistinguishable synthetic representation of a real image every time, which is possible if a discriminator is embedded within the transformers loss function (Fig. 3). As explained in \cite{goodfellow2014generative,salimans2016improved,shrivastava2016learning}, a discriminator is essentially a classifier that classifies the output of another network as real or fake. However, unlike \cite{shrivastava2016learning}, in our case the role of the discriminator is reversed\textemdash instead of enforcing the transformer to produce more realistic images the role of the discriminator is to enforce the transformer to produce synthetic images. The discriminator loss can be defined as follows

\begin{equation}\label{eq:Disc_loss}
  \begin{aligned}
 \Lno_{\D}(\gamma_d) = -\sum_{i} \log(\D_{\gamma_{d}}(\bm{x}^{'})) -\sum_{j} \log(1-\D_{\gamma_{d}}(g_j)).
  \end{aligned}
\end{equation}

\begin{figure}
\label{fig_3}
\centering
\includegraphics[width=8.5cm]{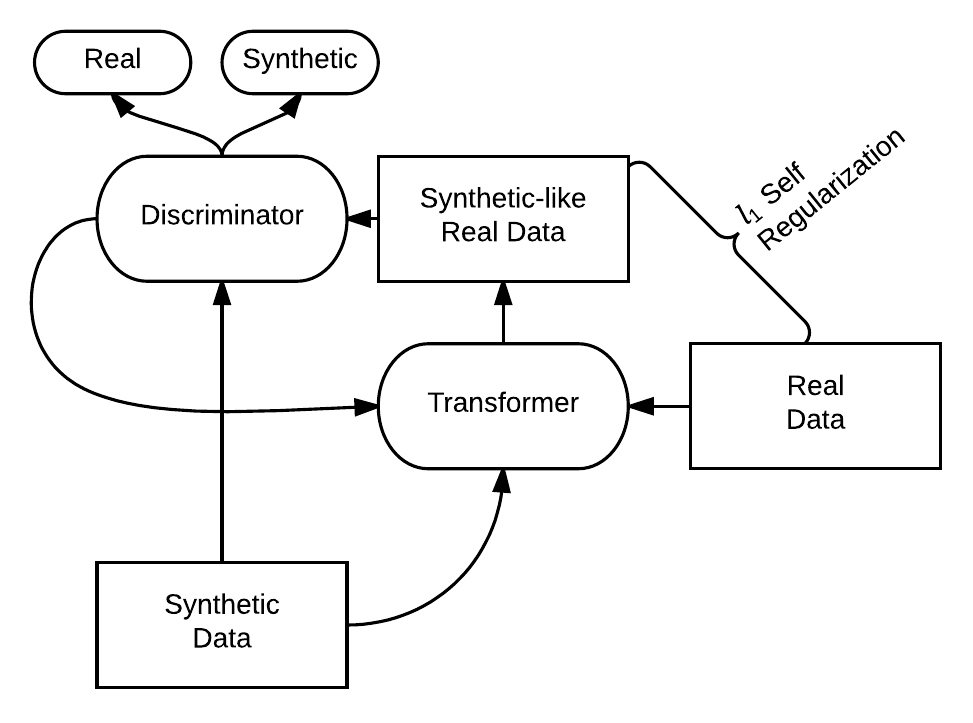}
\caption{An overview of our proposed adversarial training architecture. Real data is transformed into a synthetic-like representation using a transformer network that minimizes an adversarial loss term and a self-regularization term. The discriminator acts as a classifier to identify the image as real or synthetic and gives feedback to the transformer via the adversarial loss. The total loss essentially defines how well the discriminator is tricked into believing that the transformed real image is synthetic and how close the transformed image is to the real image.}
\end{figure}

%The loss of the discriminator is essentially composed of the probability that the transformed image is synthetic and the probability that the synthetic image is synthetic. 

This is essentially a two class classification problem with cross-entropy error where the first term represents the probability of the input being a synthetic image and the second term represents the probability of the input image being synthetic-like representation of a real image. The discriminator works on a patch level rather than the entire image to prevent artifacts. 

To train our network, we randomly sample mini-batches of synthetic images and images transformed to be synthetic by the transformer. Instead of using individual outputs of the transformer we use randomly sampled, buffered outputs and a set of randomly sampled synthetic images. This increases the stability of the adversarial training since the lack of memory can diverge the adversarial training and introduce artifacts \cite{shrivastava2016learning}. At each step the discriminator trains using this mini-batch and parameters $\gamma_d$ are updated using stochastic gradient decent (SGD). The transformer loss is then updated with the trained discriminator, the $\psi$ term in Eq. 1 can be defined as,

\begin{equation}\label{eq:Ttans}
  \begin{aligned}
\psi(\bm{x}_i,\G;{\gamma_t}) = -\log(1-\D_{\gamma_{d}}(\T_{\gamma_{t}}(\bm{x}))).
  \end{aligned}
\end{equation}
As the training shuffling between the transformer and discriminator reaches equilibrium the transformer is able to fool the discriminator every time. The loss in Eq. 3 forces the discriminator to fail to classify transformed images as synthetic-like real.  

\begin{algorithm}
\label{algo}
\caption{Adversarial training of a Transformer $x^{'}=\T_\gamma(x)$}\label{euclid}
\begin{algorithmic}[1]
\INPUT{Synthetic Data: $\bm{g}_i\in\G$, Real Data: $\bm{x}_i\in\X$, Transformer Updates/step: $n_{t}$,  Discriminator Updates/step: $n_{d}$ }
\For     {$s=1,2,3...S$}
\For     {$n_t=1,2,3...N_t$}
\State \multiline{Sample a mini-batch $\{x_1,x_2,...x_k\}$ of $k$ real images.}
\State \multiline{Update the transformer network parameters $\gamma_t$ by taking an SGD step:}
\Statex $\nabla_{\gamma_t}\frac{1}{k} \sum_{i} \psi(\bm{x}_i,\G;{\gamma_t}) + \lambda \phi(\bm{x}_i;\gamma_t)$
\EndFor
\For     {$n_d=1,2,3...N_d$}
\State \multiline{Sample a mini-batchs of $k$ synthetic images $\{g_1,g_2,...g_k\}$ and transformed real images $\{x_1,x_2,...x_k\}$.}
\State $\bm{x}^{'}_i \gets \T_\gamma(\bm{x}_i)$
\State \multiline{Update the discriminator network parameters $\gamma_d$ by taking an SGD step:}
%\State \multiline{%
 %             Considering the $k$-th edge of the $i$-th metabolite,~$x_1$ and~$y_1$ 
  %            contains the vertices of the $k$-th edge.}
\Statex $-\nabla_{\gamma_d}\frac{1}{k}\sum_{i} \log(\D_{\gamma_{d}}(\bm{x}^{'})) -\sum_{j} \log(1-\D_{\gamma_{d}}(g_j))$
\EndFor  {}
\EndFor
\OUTPUT Trained Transformer Model $\T_\gamma(x)$
%\EndProcedure
\end{algorithmic}
\end{algorithm}

\textbf{Self-Regularization:} As mentioned earlier, a key requirement for the transformer is that it should only remove patient specific data and should preserve other features such as shape. For the proof-of-concept proposed in this work, we utilize a simple, per-pixel loss term between the real image and the synthetic-like real representation of the image to penalize the transformed image from deviating significantly from the real image. The self regularization term $\phi$ can be defined as,

\begin{equation}\label{eq:self_reg}
  \begin{aligned}
  \phi(\bm{x}_i;\gamma_t) = \mid\mid\Phi(\T_{\gamma_{t}}(\bm{x}))-\Phi(\bm{x})\mid\mid_1,
  \end{aligned}
\end{equation}

where $\Phi$ represents the feature transform and $\mid\mid.\mid\mid_1$ represents the $\ell_1$ norm. 

 The transformer loss term can be rewritten as, 

\begin{equation}\label{eq:Ttans_loss2}
  \begin{aligned}
 \Lno_{\T}(\gamma_t) = -\sum_{i} \log(1-\D_{\gamma_{d}}(\T_{\gamma_{t}}(\bm{x}))) \\ + \lambda \mid\mid\Phi(\T_{\gamma_{t}}(\bm{x}))-\Phi(\bm{x})\mid\mid_1,
  \end{aligned}
\end{equation}

 In summary, the total loss measures how well the discriminator is tricked into believing that the transformed real image is synthetic, and how close the transformed image is to the real image. This overall training process has been explained in detail in Algorithm 1.

%\textbf{Stability of Adversarial Training:} 

%The goal is to use a set of synthetic images $\bm{g}_i\in\G$ to learn a transformer $\bm{x}^{'}=\T_{\gamma_{t}}(\bm{x})$ that can transform real images $\bm{x}$ to a synthetic-like representation $\bm{x}^{'}$. The transformed should be able to fool a discriminator $\D_{\gamma_{d}}$ where $\gamma_t$ and $\gamma_d$ are the learning parameters. There are three key requirements for this setup: a) The transformer output should only remove the patient specific details in the image. b) The adversarial training should not introduce artifacts in the transformed image. c) The adversarial training should be stable. The transformer loss function can be defined as,
%\section{Experiments}

\section{Depth Estimation from Monocular Endoscopy Images}

The previous sections have talked about generating synthetic medical data and adversarial training to bring real images within the domain of the synthetic data via a reverse domain adaptation pipeline. In order to evaluate the effectiveness of our proposed reverse domain adaptation pipeline we train a network from synthetically generated endoscopy data (Fig. 2) and demonstrate that it can be adapted to three different target domains. By demonstrating that distribution of the target domain can be brought closer to the source domain via adversarial training essentially showing that our depth estimation paradigm is domain independent. 

Once the synthetic data with ground truth depths is generated we use a CNN-CRF based depth estimation framework described in \cite{liu_learning_2016}. Assuming $\vec{g}\in \R^{n\times m}$ is a synthetic endoscopy image which has been divided into $p$ super-pixels and $\vec{y}=[y_1,y_2,...,y_p] \in \R$ is the depth vector for each super-pixel. In this case, the conditional probability distribution of the synthetic data can be defined as,

\begin{equation}\label{eq:Tk}
  \begin{aligned}
 {Pr(\vec{y}|\vec{x})}=\frac{exp(E(\vec{y},\vec{x}))}{\int_{-\infty}^{\infty} exp(E(\vec{y},\vec{x})) d\vec{y}}.
  \end{aligned}
\end{equation}

 where, $E$ is the energy function. In order to predict the depth of a new image we need to solve a maximum aposteriori (MAP) problem, $\widehat{\vec{y}}=\argmax_{y} {Pr(\vec{y}|\vec{x})}$. 

Let $\xi$ and $\eta$ be unary and pairwise potentials over nodes $\N$ and edges $\Sp$ of $\vec{x}$, then the energy function can be formulated as,
%\vspace{-0.4em}
\begin{equation}\label{eq:Tk}
  \begin{aligned}
 E(\vec{y},\vec{x}) = \sum_{i \in \N} \xi({y}_i,\vec{x};\vec\theta) + \sum_{(i,j) \in \Sp} \eta({y}_{i},{y}_{j},\vec{x};\vec\beta),
  \end{aligned}
\end{equation}

where, $\xi$ regresses the depth from a single superpixel and $\eta$ encourages smoothness between neighboring superpixels. The objective is to learn the two potentials in a unified CNN framework. The unary part takes a single image superpixel patch as an input and feeds it to a CNN which outputs a regressed depth of that superpixel. Based on \cite{liu_learning_2016} the unary potential can be defined as, 
%\vspace{-1.3em}

\begin{equation}\label{eq:Tk}
  \begin{aligned}
 \xi(y_{i},\vec{x};\vec\theta)= - (y_i-h_i(\vec{\theta}))^2
  \end{aligned}
\end{equation}

 where $h_i$ is the regressed depth of superpixel and $\theta$ represents CNN parameters. 
 %The network operates on a $224\times224$ superpixel patch level and is composed of 5 convolutional and 4 fully connected layers. 

 The pairwise potential function is based on standard CRF vertex and edge feature functions studied extensively in \cite{qin_global_2009} and other works. Let $\vec{\beta}$ be the network parameters and $\vec{S}$ be the similarity matrix where ${S}_{i,j}^k$ represents a similarity metric between ithe $i^{th}$ and $j^{th}$ superpixel. Since inverse of intensity is a very valuable cue for depth estimation in endoscopy settings, we use intensity difference and greyscale histogram as pairwise similarities expressed in the general $\ell_2$ form. The pairwise potential can then be defined as,

\vspace{-1.0em}

\begin{equation}\label{eq:Tk}
  \begin{aligned}
 \eta(y_{i},y_{j};\vec\beta)= - \frac{1}{2}\sum_{k=1}^{K}\beta_{k}S_{i,j}^{k}(y_i-y_j)^2. 
  \end{aligned}
\end{equation}

The overall energy function can now be written as,

\vspace{-1.0em}

\begin{equation}\label{eq:Tk}
%\vspace{-0.7em}
  \begin{aligned}
 E = - \sum_{i \in \N} (y_i-h_i(\vec{\theta}))^2  - \frac{1}{2}\sum_{(i,j) \in \Sp}\sum_{k=1}^{K}\beta_{k}S_{i,j}^{k}(y_i-y_j)^2.
  \end{aligned}
\end{equation}

For training the negative log likelihood of the probability density function which can be calculated from Eq. 6 is minimized with respect to the two learning parameters. Two regularization terms are added to the objective function to penalize heavily weighted vectors $(\lambda_\theta,\lambda_\beta)$. Assuming $N$ is the number of images in the training data,

\vspace{-1.0em}

\begin{equation}\label{eq:Tk}
  \begin{aligned}
  \min_{\theta,\beta \geq 0} {-\sum_{1}^{N}\log Pr(\vec{y}|\vec{x};\vec{\theta},\vec{\beta}})+ \frac{\lambda_\theta}{2} \norm{\theta}_2^2+ \frac{\lambda_\beta}{2} \norm{\beta}_2^2.
  \end{aligned}
\end{equation}

The optimization problem is solved using stochastic gradient decent-based back propagation.

\section{Experiments}

\subsection{Evaluation Datasets} 

We use three kinds of datasets in our quantitative and qualitative study of the proposed methods. Since there are no publicly available endoscopy datasets with ground true depth, we generate two kinds of datasets for quantitative evaluation: a) images from a virtual endoscope in a colon phantom, and b) CT-registered optical endoscopy data collected from a real porcine colon (Fig. 4). We also use publicly-available human colonoscopy images to qualitatively assess if intuitive depth maps can be generated from real endoscopy videos.

\textbf{Colon Phantom Data:} The colon phantom data is generated from a CT-reconstructed model of a colon phantom molded from a real colon (Chamberlain Group Colonoscopy Trainer, SKU 2003\footnote{\fm{https://www.thecgroup.com/product/colonoscopy-trainer-2003/}}). A virtual endoscope is used to render images from a variety of endoscopy images with corresponding ground truth from the CT-reconstructed model. 2,160 images are generated via this procedure and are used for evaluation (Fig. 4).

\textbf{Real Porcine Colon Data:} Real endoscopy images were recorded from a pig colon fixed to a scaffold. A 3D model of the scaffold was then acquired with a CT measurement, and ground truth depth was generated for each real endoscopy image by registering virtual endoscopy views from the CT and optical endoscopy views from an endoscope (Fig. 4). 1,400 images with corresponding ground truth depth are generated using this procedure and are used for evaluation.

\textbf{Real Endoscopy Data:} We also evaluate our networks on publicly available endoscopy data\footnote{\fm{https://polyp.grand-challenge.org/databases/}} \cite{bernal2015wm,tajbakhsh2016automated}. However, these datasets do not have ground true depth and can only be used for qualitative evaluations.

\begin{figure}
\label{fig_4}
\centering
\includegraphics[width=8.5cm]{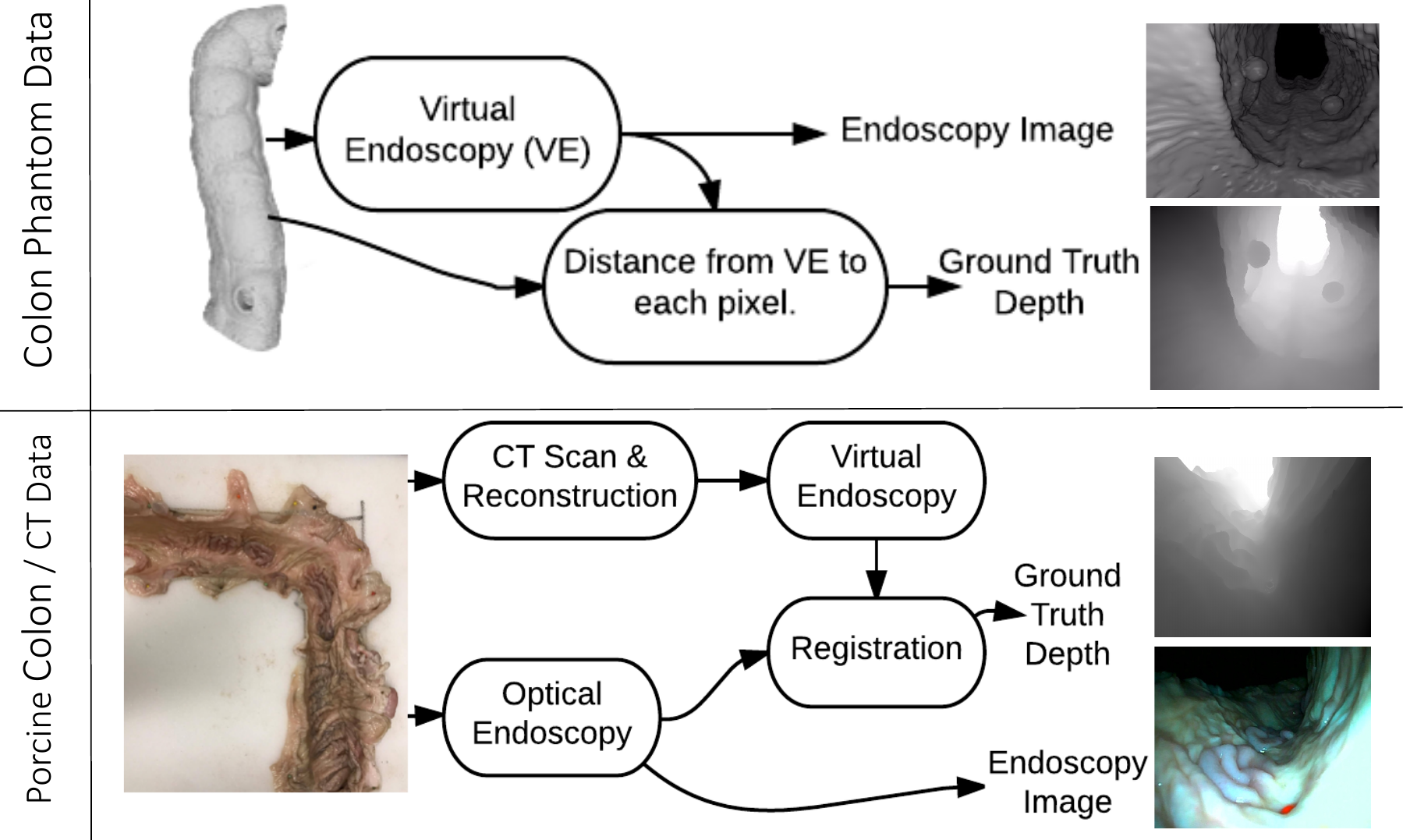}
%\caption{The data collection pipeline for colon phantom data and porcine colon data. The colon phantom data is collected from a 3D rendered colon phantom using a virtual endoscope. The ground true depth can be calculated using the 3D model. The porcine colon/CT data is collected by  imaging a porcine colon mounted on a scaffold using an optical endoscope and reconstructing a 3D model of the scaffold using CT. The optical endoscopy and CT views are then registered to get the ground true depth. }
\caption{The image collection and generation pipeline for colon phantom data and porcine colon data. The colon phantom data is collected from a 3D rendered colon phantom using a virtual endoscope. The ground truth depth is calculated using the 3D model. The porcine colon data is collected by  imaging a porcine colon mounted on a scaffold using an optical endoscope and reconstructing a 3D model of the colon from CT measurements. The optical endoscopy and CT views are then registered to get the ground truth depth maps.}
\vspace{-1.4em}
\end{figure}

%2,000 endoscopy images with their ground true depths are generated using. A

\subsection{Depth Estimation Network Trained on Synthetic Images}

\textbf{Implementation Details} \\
%The architecture used for training an endoscopy depth estimation network included training the unary and a pairwise parts of a CRF in a unified framework presented in \cite{liu_learning_2016}. The unary part is composed of a fully convolutional network which generates convolution maps which are fed into a superpixel poopling layer followed by three fully connected layers. The pairwise part operates on a super-pixel level and is composed of a single fully connected layer. This setup was implemented using VLFeat Mat-ConvNet using MATLAB 2017a and CUDA 8.0. The training data was prepared by over-segmenting each virtual endoscopy image into superpixels and corresponding ground truth depth was assigned to each superpixel. Synthetic endoscopy data and its corresponding ground truth depth generated according to the synthetic data generation pipeline presented in Section 3. The generated data was randomized to prevent the network from learning too many similar features quickly. $55\%$ of the data was used for training and $40\%$ for validation and $5\%$ for testing. Training was done using K80 GPUs. Momentum was set at 0.9 as suggested in \cite{liu_learning_2016} and both weight decay parameters in Eq. 11 $(\lambda_{\theta},\lambda_{\beta})$ were set to 0.0007. The learning rate was initialized at 0.00001 and decrease by 20\% every 20 epoches. These parameters were tuned to achieve best results. A total of 300 epochs were run and the epochs with least $\log10$ error were selected to avoid the selection of an over-fitted model. 

The architecture used for training an endoscopy depth estimation network includes training the unary and a pairwise parts of a CRF in a unified framework presented in \cite{liu_learning_2016}. The unary part is composed of a fully convolutional network which generates convolution maps that are fed into a superpixel pooling layer followed by three fully connected layers. The pairwise part operates on a superpixel level and is composed of a single fully connected layer. This setup was implemented using VLFeat Mat-ConvNet\footnote{\fm{http://www.vlfeat.org/matconvnet/}} using MATLAB 2017a and CUDA 8.0. The training data was prepared by over-segmenting each virtual endoscopy image into superpixels and corresponding ground truth depth were assigned to each superpixel. Synthetic endoscopy data and its corresponding ground truth depth was generated according to the synthetic data generation pipeline presented in Section 3. The generated data was randomized to prevent the network from learning too many similar features quickly. $55\%$ of the data was used for training and $40\%$ for validation and $5\%$ for testing. Training was done using K80 GPUs. Momentum was set at 0.9 as suggested in \cite{liu_learning_2016} and both weight decay parameters in Eq. 11 $(\lambda_{\theta},\lambda_{\beta})$ were set to 0.0007. The learning rate was initialized at 0.00001 and decrease by 20\% every 20 epochs. These parameters were tuned to achieve best results. A total of 300 epochs were run and the epochs with least $\log10$ error were selected to avoid the selection of an over-fitted model. 

\subsection{Adversarial Training for Reverse Domain Adaptation}

%\subsubsection{Implementation}

\textbf{Implementation Details} \\
Since the depth estimation network was trained solely on synthetic data all test images need to have a synthetic-like representation for the depth estimation to perform effectively. A transformer network was trained using the reverse domain adaption paradigm presented in Section 4.

The transformer and discriminator networks were implemented using tensorflow. The synthetic and real endoscopy images were down-sampled to a pixel size of $244\times244$ for computational efficiency. The real images were also converted to grayscale. The training between the transformer and the discriminator proceeds alternatively.

%The transformer network was a standard residual network ResNet \cite{he2016deep}. The setup is similar to \cite{shrivastava2016learning} but for refining real data to be synthetic rather than the other way around. The input image of size $244\times244$ is convolved with a filter of $7\times7$ that outputs $64$ feature maps which is then passed to 10 ResNet blocks followed by a $1\times1$ convolution layer resulting in one feature map.  The transformer is first trained with only the self-regularization term for the first 800 steps and the discriminator for 200 steps. The discriminator network is a standard classifier with five convolution layers, two max-pooling layers and softmax. More implementation details are given in the supplement.

\begin{figure}
\label{fig_5}
\centering
\includegraphics[width=7.5cm]{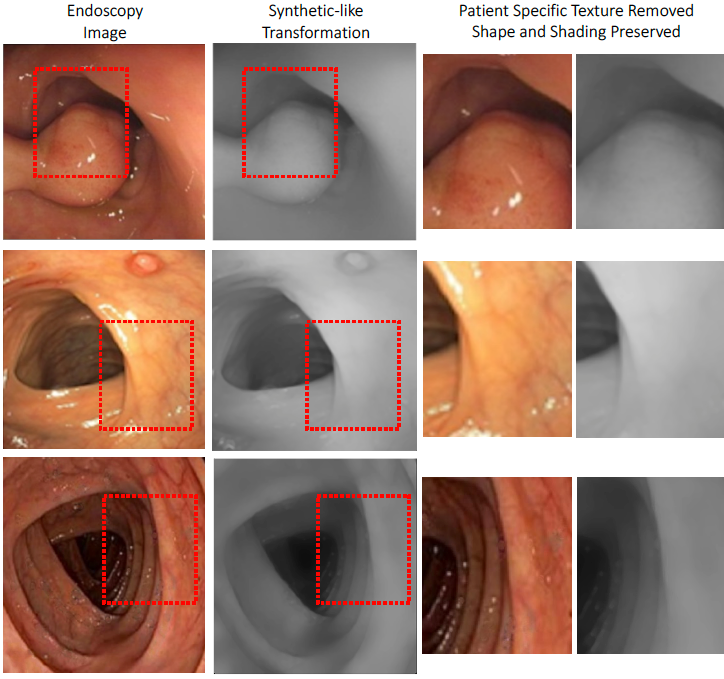}
\caption{Examples of real endoscopy images transformed to their synthetic-like representations. Patient-specific texture is clearly removed during the transformation.}
\vspace{-1.2em}
\end{figure}

\begin{figure}
\label{fig_6}
\centering
\includegraphics[width=6.5cm]{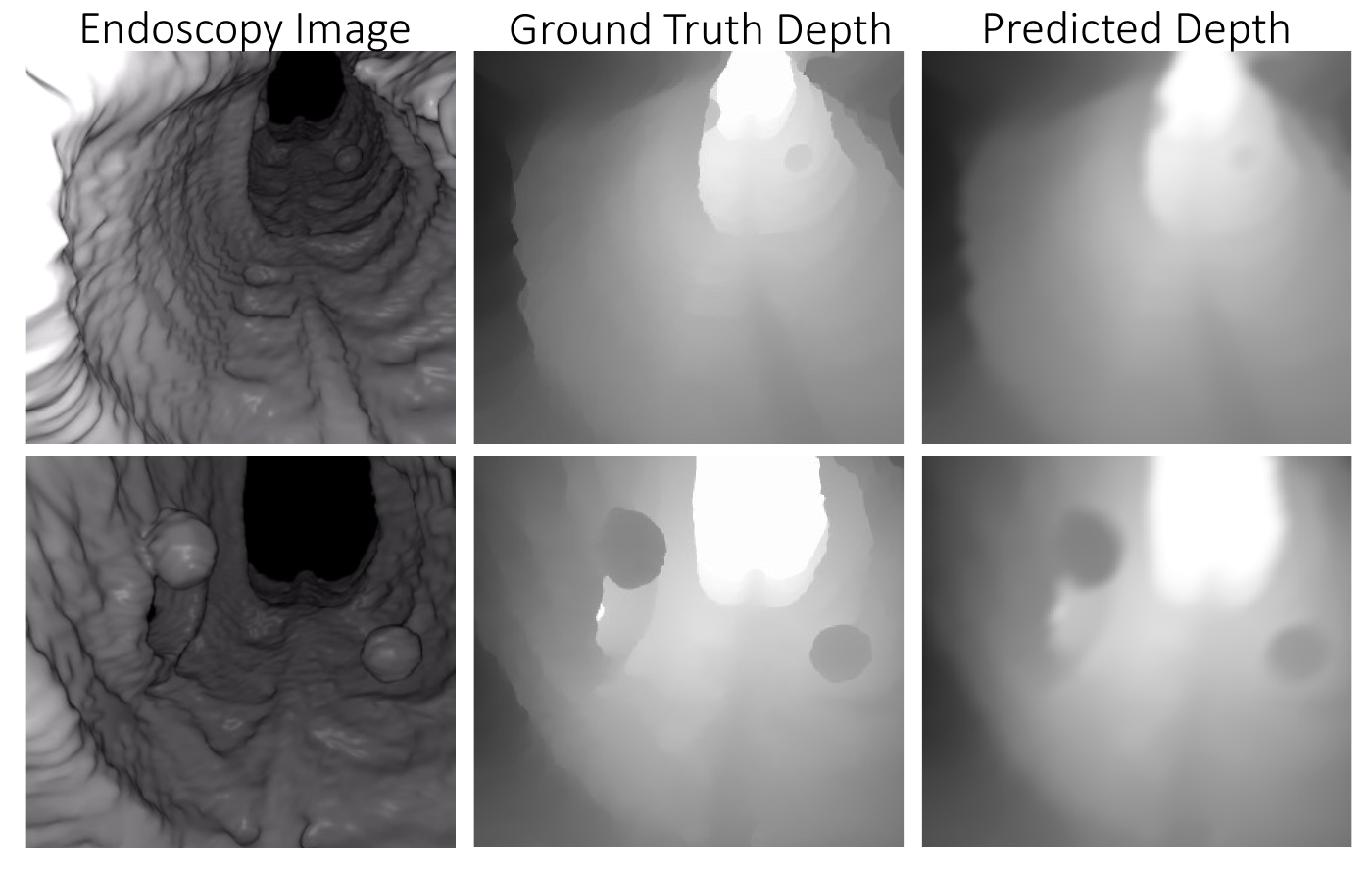}
\caption{Examples of rendered images, corresponding ground truth depth, and depth estimates from a colon phantom.}
\vspace{-0.3em}
\end{figure}

The transformer network was a standard residual network (ResNet) \cite{he2016deep}. This is similar to \cite{shrivastava2016learning}, but for refining real data to be synthetic rather than the other way around. An input image of size $244\times244$ is convolved with a filter of $7\times7$ that outputs $64$ feature maps which are then passed to 10 ResNet blocks followed by a $1\times1$ convolution layer resulting in one feature map. The transformer is first trained with only the self-regularization term for the first 800 steps and the discriminator for 200 steps. The discriminator network is a standard classifier with five convolution layers, two max-pooling layers and softmax.

\begin{table}
\begin{center}
\begin{tabular}{|l|l|c|c|c|}
\hline
Test Dataset & NRMSE & HD & SSIM  \\
\hline\hline
Colon Phantom  & 0.38& 0.36 & 0.52\\
Trans. Colon Phantom  & \textbf{0.23}& \textbf{0.23} & \textbf{0.77}\\
\hline\hline
Real Porcine Colon &  0.61 & 0.58 &0.33\\
Trans. Real Porcine Colon   & \textbf{0.32} & \textbf{0.30}& \textbf{0.59}\\
\hline
\end{tabular}
\end{center}
\caption{A comparison between depth estimated from raw images and domain adapted images via our transformer network.}
\end{table}

\begin{table}
\begin{center}
\begin{tabular}{|l|l|c|c|c|}
\hline
Method & NRMSE & HD & SSIM  \\
\hline\hline
DiL (No Texture)\cite{nadeem2016computer}  & 0.57 & 0.56& 0.35\\
DiL (Texture 1)\cite{nadeem2016computer}  & 0.49 & 0.44& 0.31\\
DiL (Texture 2)\cite{nadeem2016computer} & 0.43 & 0.43& 0.30\\
DiL (Average)   & 0.50 & 0.48& 0.32\\
Ours (No Texture)&  0.19& 0.18&0.81  \\
Ours (Phantom)&  0.23& 0.23 & 0.77\\
Ours (Porcine)&  0.32 & 0.30 &0.59\\
Ours (Average)& \textbf{0.25}& \textbf{0.24}&\textbf{0.72}\\
\hline
\end{tabular}
\end{center}
\caption{Results of our method as compared to the state-of-the-art endoscopy depth estimation method.}
\end{table}

\begin{figure}
\label{fig_7}
\centering
\includegraphics[width=8.0cm]{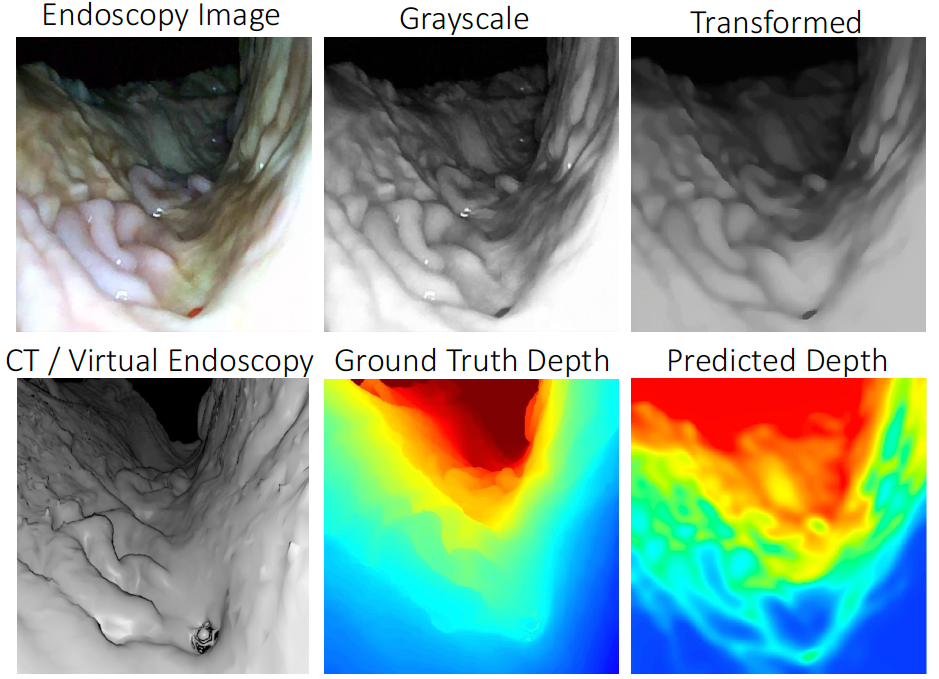}
\caption{Depth estimates from porcine colon data. The optical endoscopy image is converted to grayscale and transformed to its synthetic-like representation using our transformer network. The optical endoscopy view is registered to it's corresponding CT view to obtain ground truth depth.}
\vspace{-1.0em}
\end{figure}

\subsection{Results}

\begin{figure*}
\label{fig_8}
\centering
\includegraphics[width=\textwidth]{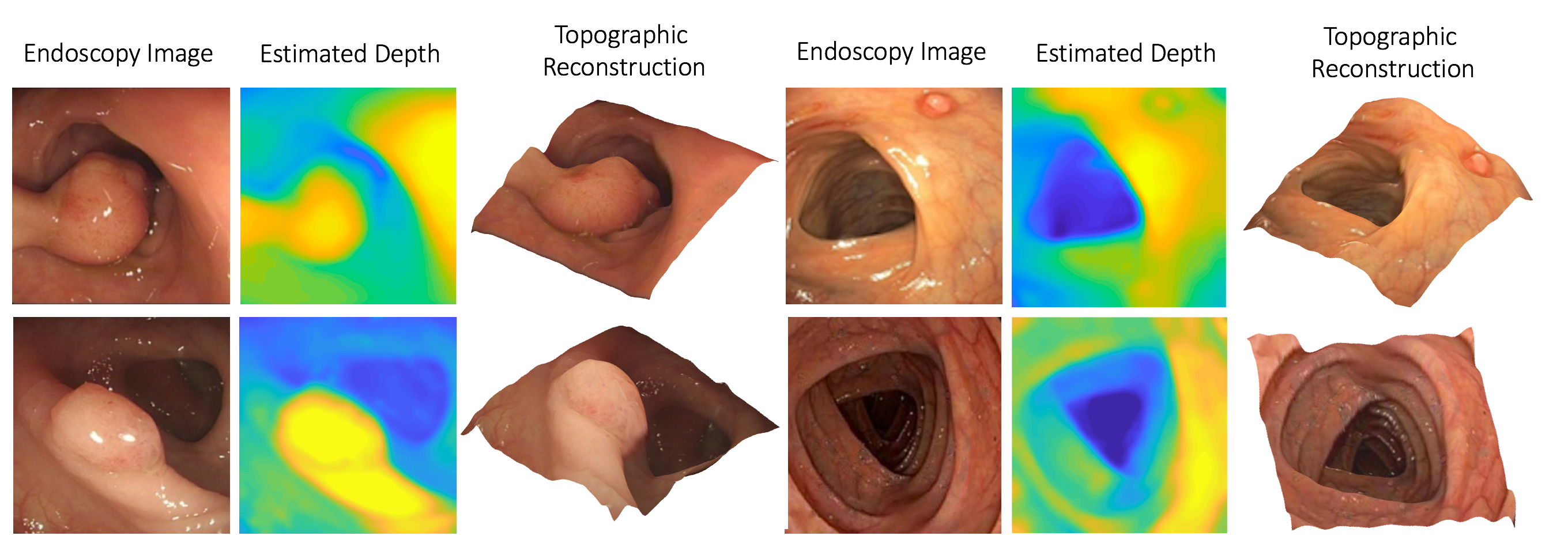}
\caption{Depth estimates and topographical reconstructions from monocular endoscopy images. Each endoscopy image is transformed to its synthetic-like representation as shown in Fig. 5 and is fed into our depth estimation network. The depth is then used to reconstruct the surface topography.}
\vspace{-0.6em}
\end{figure*}

\textbf{Transformer Network:} Fig. 5 shows examples of real endoscopy images transformed to their synthetic-like representations. It can clearly be seen that the patient specific information has been removed and clinically-relevant features have been preserved for depth estimation and polyp identification. A close-up of the images show that the vasculature has been removed while preserving the shape information. In the next subsections we demonstrate that the depth estimation network trained on synthetic data performs significantly better with images transformed to their synthetic like representations.

%Fig. 5 shows examples of real endoscopy images transformed to their synthetic-like representations. It can clearly be seen that the patient specific information has been removed and clinically-relevant features have been preserved. A close-up of the images show that the vasculature has been removed while preserving the intensity information. In the next subsections we demonstrate that the depth estimation network trained on synthetic data performs significantly better with images transformed to their synthetic like representations.

\textbf{Depth Evaluation Metrics:} We compare our depth estimates to corresponding ground truth values based on three metrics which have been used in previous endoscopy depth estimation work: 

\begin{itemize}
\item Normalized root mean square error (NRMSE): NRMSE = $\frac{\sqrt{\frac{\sum_i{(x_i-y_i)}}{n}}}{(x_{max}-x_{min})}$, is a normalized RMS error for comparative analysis across datasets. A lower RMSE indicates the data being compared is more similar.

\item Hausdorff distance (HD): HD calculates the greatest of all the distances from a point in the ground truth data to the closest point in the calculated data \cite{huttenlocher1993comparing}. It can be calculated as $H(x,y)=\max(\vec{h}(x,y),\vec{h}(y,x))$ where $\vec{h}=\max_{a} \min_{b}\norm{a-b}$. A lower HD indicates the two datasets being compared are more similar.

\item Structural Similarly Index (SSIM): The SSIM is an image assessment index calculated on the basis of luminance, contrast and structure. The SSIM in this paper is calculated according to the definition proposed in \cite{wang2004image}. This index is between $-1$ and $1$ with 1 indicating identical images.
\end{itemize}

\textbf{Quantitative Results}

%Table 1 shows a comparative analysis of depth estimation results from colon phantom and real porcine colon data with and without transformation. It can clearly be seen that without transforming the data to it's synthetic-like representation the depth estimation network trained on synthetic data behaves poorly. Without transformation the results with the colon phantom data are slight better as compared to porcine colon data because the colon phantom has less patient specific texture. Fig. 6 shows results for the colon phantom and Fig. 7 shows depth estimation results from the porcine colon. There is a \textbf{88\% improvement} in the SSIM for the porcine colon data and a \textbf{48\% improvemen}t for the colon phantom data by transforming the input data using our proposed paradigm.

Table 1 compares depth estimation results from colon phantom and real porcine colon data with and without domain transformation. It can clearly be seen that depth estimation is improved by domain transformation. As expected, the improvement in depth estimation that domain transformation provides is marginal in the colon phantom data, which has homogenous material properties, and more significant in real porcine tissue, which has natural biological variation in mucosal texture. There is a \textbf{88\% improvement} in the SSIM for the porcine colon data and a \textbf{48\% improvement} for the colon phantom data by transforming the input data using our proposed paradigm. Fig. 6 and 7 show representative depth estimation results for the colon phantom and porcine colon, respectively. 

%\textbf{Real Porcine Colon Data - Quantitative Results}

\textbf{Comparative Analysis}

%Due to the lack of available ground truth endoscopy depth data there is currently only one learning based monocular colonoscopy depth estimation study by Nadeem {et al.} \cite{nadeem2016computer}. They use dictionary learning (DiL) and use CT colonoscopy data for training. However, unlike our data their data does not follow optically correct inverse square intensity fall off. Table 2 shows a comparative analysis with their results. We demonstrate that our depth estimation approach is significantly better than their method and improves the SSIM by 128\%. 

Due to the lack of available ground truth endoscopy depth data there is currently only one learning based monocular colonoscopy depth estimation study by Nadeem \textit{et al.} \cite{nadeem2016computer}. They implement dictionary learning (DiL) and use CT colonoscopy data for training. However, unlike our data, their data does not follow optically-correct inverse square intensity fall off, which we expect to be a significant cue for absolute depth. Table 2 shows a comparative analysis of their results compared to those from our approach. We demonstrate that our depth estimation is significantly better than their method and \textbf{improves the SSIM by 125\%}. 

\textbf{Qualitative Results}\\
For the purposes of demonstration, we also show that it is possible to estimate depth from real human endoscopy data. Fig. 8 shows monocular endoscopy images, their estimated depth, and corresponding topographic reconstructions. Topographical reconstructions are reconstructed by overlaying depth on a 3D manifold. These depth estimates are qualitative and there is no corresponding ground truth depth available.

\section{Conclusions and Future Work}

In this paper, we propose a novel reverse domain adaptation method that transforms real medical images into useful synthetic representations while preserving clinically relevant features. We validated this method in the task of monocular depth estimation for endoscopy images, in which we first learned depth from a large synthetic dataset, and then demonstrated an 88\% and 48\% improvement in predicting depth from synthetic-like domain-adapted images over raw images for both a real porcine colon and a colon phantom respectively. Future work will focus on using the proposed reverse domain adaption paradigm for other medical imaging modalities and on using the predicted depth for improving automated polyp segmentation and classification.

{\small
\bibliographystyle{ieee}
\bibliography{egpaper_for_review.bbl}
}

\end{document}